\documentclass[letterpaper]{article} 
\usepackage[preprint]{aaai2027}  
\usepackage[hyphens]{url}  
\usepackage{graphicx} 
\usepackage{natbib}  
\usepackage{caption} 
\usepackage{algorithm}
\usepackage{algorithmic}

\usepackage{booktabs}

\usepackage{amsmath}
\usepackage{amssymb}

\DeclareMathOperator*{\argmax}{arg\,max}

\title{Think in Sets for Streaming Video Token Compression}

\author{Moxu Duan\textsuperscript{\rm 1},
Jingwen Fu\textsuperscript{\rm 2}\corresponding,
Yuwang Wang\textsuperscript{\rm 3}\corresponding}
\affiliations{
\textsuperscript{\rm 1}Beijing Jiaotong University, \textsuperscript{\rm 2}Zhongguancun Academy, \textsuperscript{\rm 3}Tsinghua University

24271096@bjtu.edu.cn,
fujingwen@bza.edu.cn,
wang-yuwang@tsinghua.edu.cn
}

\begin{document}

\maketitle

\begin{abstract}
Streaming VideoLLMs process frames causally while visual tokens grow continuously, making compression essential for controlling prefilling latency and memory. Existing training-free methods independently rank tokens, ignoring marginal-gain interactions among retained tokens. We argue that streaming video token compression should instead be formulated as set selection, where each candidate is valued by what it adds beyond the tokens already retained. Unlike existing set-wise methods designed for offline tasks, streaming makes causal, frame-by-frame pruning decisions, so modeling cross-frame interactions requires an explicit historical reference. This creates a  \emph{reference-set dilemma}: the reference must adequately represent previously conveyed content while remaining bounded for real-time inference. We introduce NovaCov, to our knowledge the first training-free, plug-and-play set-wise token compressor designed for streaming video. NovaCov maintains a capacity-bounded, recency-weighted Historical Reference Bank and optimizes a dual-branch submodular coverage objective that preserves representative current-frame content while prioritizing information insufficiently covered by history. Both branches are facility-location functions, so greedy selection retains the classical $(1-1/e)$ approximation guarantee. Across streaming and offline benchmarks, NovaCov outperforms existing training-free compression methods, retaining 99.6\% of ReKV accuracy while reducing LLM prefilling latency by 46\%.
\end{abstract}

\section{Introduction}

\begin{figure}[t]
\centering
\includegraphics[width=\columnwidth]{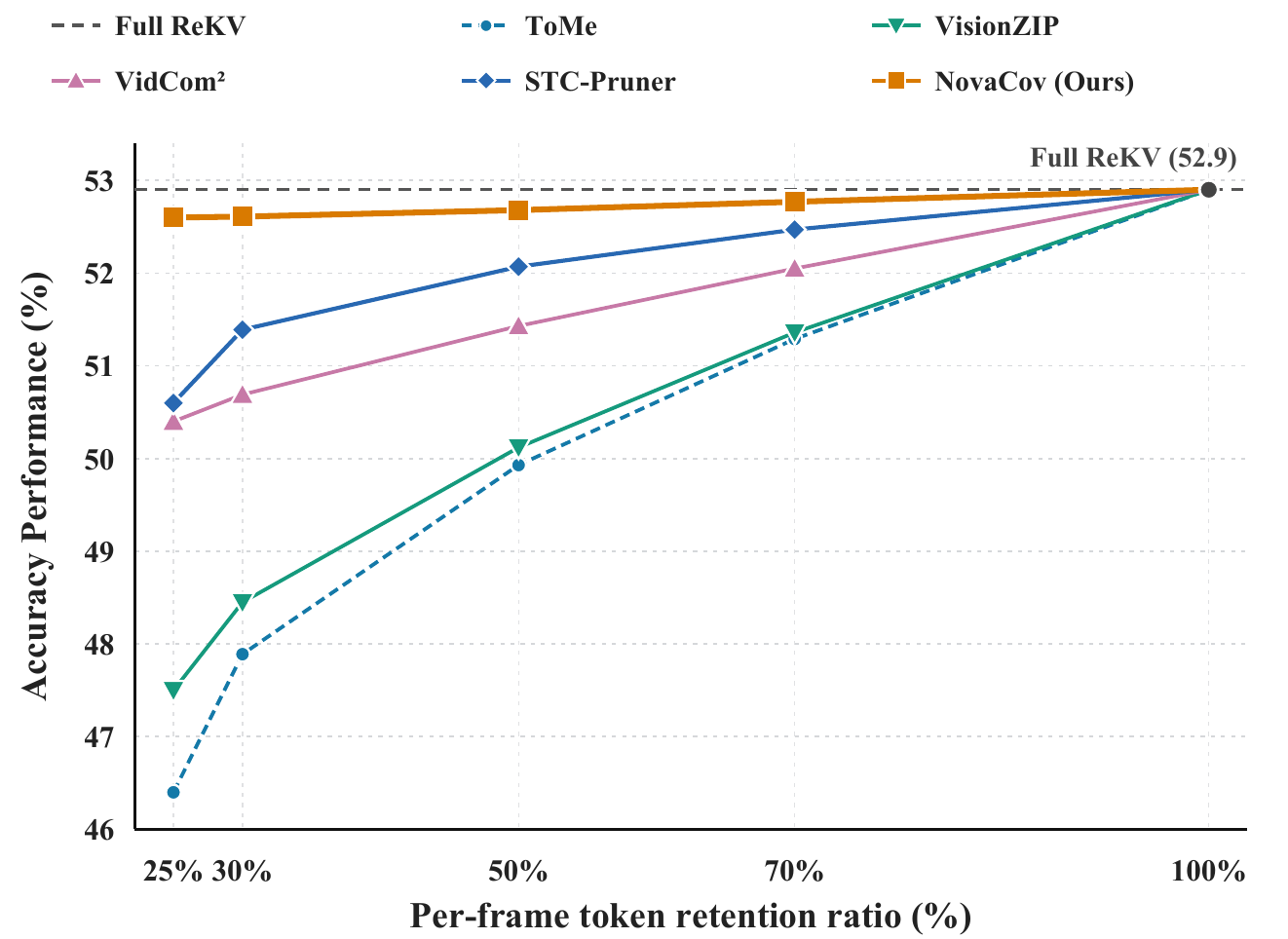}
\caption{Accuracy–budget trade-off on OVO-Bench under different per-frame token-retention ratios. Among representative training-free token compression methods, NovaCov consistently achieves the highest accuracy across all compression levels.}
\label{fig:teaser}
\end{figure}

Streaming video understanding is an increasingly important capability of video large language models (VideoLLMs)~\citep{chen2025longvila,wang2025internvideo25,wang2026videoitg,zhang2024videoinstruction,zhang2025videollama3}, enabling multimodal understanding and reasoning over continuously arriving visual observations~\citep{li2024mvbench,fu2025videomme,zhou2025mlvu}. Its practical deployment, however, is constrained by the cost of processing dense visual tokens~\citep{chen2025streamingtom}. Unlike offline video, where the complete recording and its temporal boundary are known, streaming video requires the model to process each frame causally and remain ready for a query at any moment. The model therefore has no access to future frames or future questions, while visual tokens continue to accumulate as the stream progresses~\citep{wei2025streamvln,lin2026streamingbench}. This accumulation increases prefilling latency and KV-cache memory, making visual-token compression before the LLM essential for practical streaming inference~\citep{ning2025livevlm}. 

What remains unclear is not whether tokens should be compressed, but \emph{how they should be selected}. Existing training-free streaming methods~\citep{yao2025timechatonline,chen2025streamingtom,wang2025stc,xie2026fluxmem} typically assign each token an independent score based on signals such as spatial saliency, inter-frame difference, or distance to a running anchor, and retain the top-ranked tokens. Regardless of
the scoring signal, these methods share the same underlying
structure: the value of a token is treated as an intrinsic property, independent of the other tokens retained. Several individually important tokens may therefore represent nearly identical content, repeatedly spending a limited budget on the same object, region, or background.

\begin{figure*}[t]
\centering
\includegraphics[width=\textwidth]{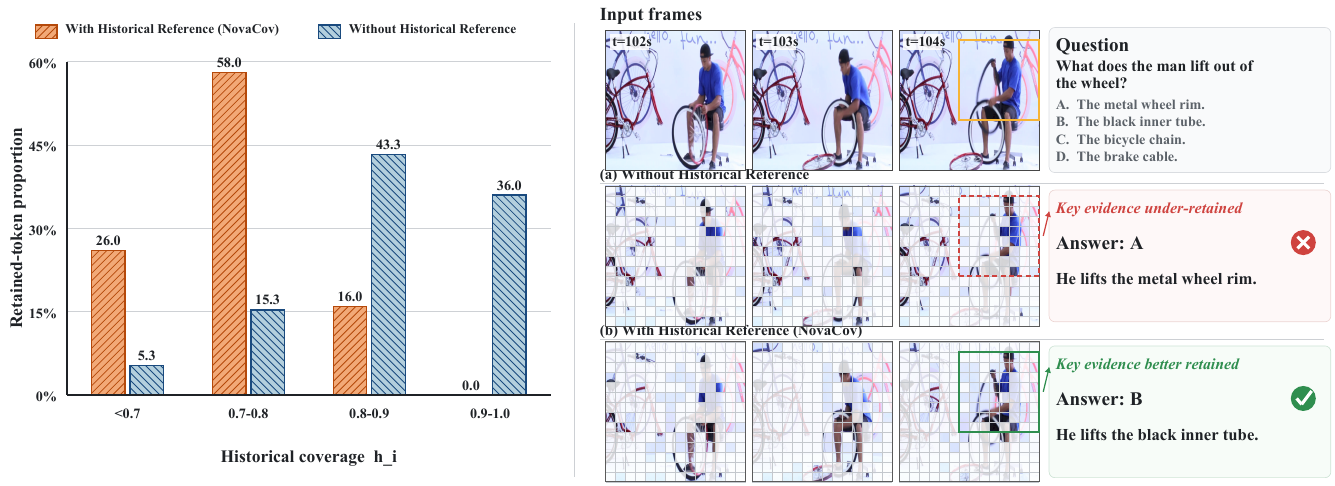}
\caption{Visual diagnosis of the need for an explicit historical reference in streaming set-wise selection.
Left: historical coverage $h_i$ of retained tokens, computed post hoc using the same Historical Reference Bank. NovaCov assigns 84.0\% of the budget to tokens with $h_i<0.8$, versus 20.6\% without historical reference.
Right: an OVO-Bench example where selection without historical reference misses key evidence, while NovaCov preserves it and answers correctly.}
\label{fig:ReferenceSet}
\end{figure*}

We argue that this token-wise formulation is fundamentally mismatched with token compression. Compression is a subset-selection problem: downstream performance depends on the collective information carried by the retained tokens, not the isolated score of each token. A \emph{set-wise} objective instead values a candidate by its marginal contribution relative to a reference set representing the content already accounted for, naturally suppressing duplicates and favoring complementary information. Recent offline methods adopt this perspective through submodular coverage maximization~\citep{dong2025mmtok,cho2026floc} or column subset selection~\citep{lee2026moving}, consistently improving over independent token ranking.

In streaming scenarios, consecutive frames often contain the same background, objects, and scene layout, while only a small portion of the visual content changes~\citep{yao2025timechatonline}. Set-wise selection can directly address this problem by asking whether a candidate token adds anything beyond
what is already retained. However, transferring it to streaming is not straightforward. Offline set-wise methods perform one-shot selection over a fully visible video, allowing selections across frames to interact through their marginal gains. Streaming instead compresses each frame upon arrival, with future frames unavailable and past selections already forwarded to the model. Applying an offline set-wise objective directly to streaming decomposes global selection into independent per-frame selections over current-frame candidates, leaving cross-frame redundancy unaccounted for. Fig.~\ref{fig:ReferenceSet} visualizes the resulting loss of cross-frame interaction. As a result, streaming set-wise selection requires an explicit historical reference to restore marginal interactions among tokens retained across frames. However, under the causal and real-time constraints of streaming, this historical reference faces a \textbf{reference-set dilemma}: On the one hand, \emph{causal completeness} requires the reference set to represent as much as possible of the information already conveyed to the model; On the other hand, \emph{real-time efficiency} requires the reference set to remain bounded.

This dilemma motivates the central question of this work: \textit{How to summarize the visual content that has already reached the model well enough to judge whether a new token contributes?} To solve this problem, we introduce \textbf{NovaCov}, to our knowledge the first training-free, set-wise token compressor for streaming video. NovaCov is a plug-and-play method built on two complementary designs. \textbf{First}, it maintains a \emph{Historical Reference Bank}, a capacity-bounded and recency-weighted summary of previously retained content. An online match-or-insert update consolidates redundant observations, while temporal decay gradually removes stale information. The bank therefore remains fixed in size and keeps the per-frame cost independent of stream length, while staying aligned with the visual context currently represented by the model. \textbf{Second}, NovaCov optimizes a dual-branch submodular coverage objective. The current-frame branch preserves the spatial content of the incoming frame, while the historical-novelty branch rewards only coverage that is not already supplied by the reference bank. A token is therefore valued not by an isolated importance score, but by what it adds to the retained set and what the model still lacks. Both branches are facility-location functions defined over different similarity matrices, so their weighted combination remains monotone submodular and greedy selection retains the classical $(1-1/e)$ approximation guarantee. We evaluate NovaCov on both streaming and offline video-understanding benchmarks. As shown in Fig.~\ref{fig:teaser}, NovaCov consistently achieves a stronger accuracy--budget trade-off across compression levels. It outperforms existing training-free compression methods in both settings, retains 99.6\% of the uncompressed ReKV~\citep{di2025rekv} accuracy, and reduces language-model prefilling latency by 46\%.

Overall, this paper makes the following contributions:
\begin{itemize}
\item We formulate streaming video token compression as a set-selection problem requiring an explicit historical reference, and identify the \emph{reference-set dilemma}: set-wise selection requires a sufficiently complete representation of historical context, whereas real-time streaming requires this representation to remain bounded.
\item We propose NovaCov, which resolves this dilemma through a capacity-bounded, recency-weighted Historical Reference Bank and a dual-branch submodular objective that jointly preserves current-frame content and prioritizes information absent from history. And we show that both objective branches are facility-location functions, preserving monotone submodularity and the classical $(1-1/e)$ approximation guarantee under greedy selection.
\item Extensive experiments demonstrate state-of-the-art performance among training-free methods on streaming and offline benchmarks, with consistently stronger accuracy--efficiency trade-offs.
\end{itemize}

\section{Related Work}
 
\paragraph{Streaming Video Understanding.}
Streaming video understanding requires a pretrained VideoLLM~\citep{li2024llavaonevision,bai2025qwen25vl,wang2026videoitg} to process continuously arriving frames while remaining ready to answer questions posed at any moment. A line of offline-to-online frameworks adapts existing VideoLLMs to this causal setting without retraining. ReKV~\citep{di2025rekv} maintains a frame-wise KV cache and retrieves relevant history for online question answering, while others reorganize the streaming context or bound the active memory through query-agnostic cache compression, eviction, or retrieval~\citep{liu2024streamchat,ning2025livevlm,yang2025streammem,kim2026infinipot}. Every arriving frame nonetheless still yields dense visual tokens that must all be prefilled, leaving causal, query-agnostic compression before the LLM as the missing piece.

\paragraph{Token Compression for Video LLMs.}
Existing training-free methods fall broadly into two categories.
 
\textbf{(i)} \emph{Token-wise Methods} assign each token an independent score. Offline designs either merge semantically or spatiotemporally redundant representations~\citep{bolya2023tome,ren2023testa,shao2025holitom}, or rank tokens by attention, text relevance, visual saliency, or temporal redundancy~\citep{chen2024fastv,tao2025dycoke,zhang2025sparsevlm,yang2025visionzip,shen2025fastvid}. Streaming-specific methods further exploit causal temporal cues, through adjacent-frame differencing, joint consistency and saliency scoring, distances to spatial and temporal anchors, or decay with temporal distance~\citep{wang2025stc,chen2025streamingtom,xie2026fluxmem}. The retained subset is not evaluated jointly, leaving redundancy among selected tokens unaddressed.
 
\textbf{(ii)} \emph{Set-wise Methods} formulate token compression explicitly as a set-level subset-selection problem, valuing a token by its marginal contribution to what is already selected. Arising from different mathematical tools, they converge on this same view: FLoC~\citep{cho2026floc} maximizes facility-location coverage over a fully visible video or temporal block; MMTok~\citep{dong2025mmtok} covers both textual and visual targets simultaneously through multimodal maximum coverage; and SPARE~\citep{lee2026moving} casts token pruning as a column subset selection problem, retaining the subspace spanned by the full visual feature matrix. These methods, however, assume a static or fully visible input and none is designed for a causal stream. To our knowledge, NovaCov is the first set-wise compressor built for streaming video.


\begin{figure*}[t]
\centering
\includegraphics[width=\textwidth]{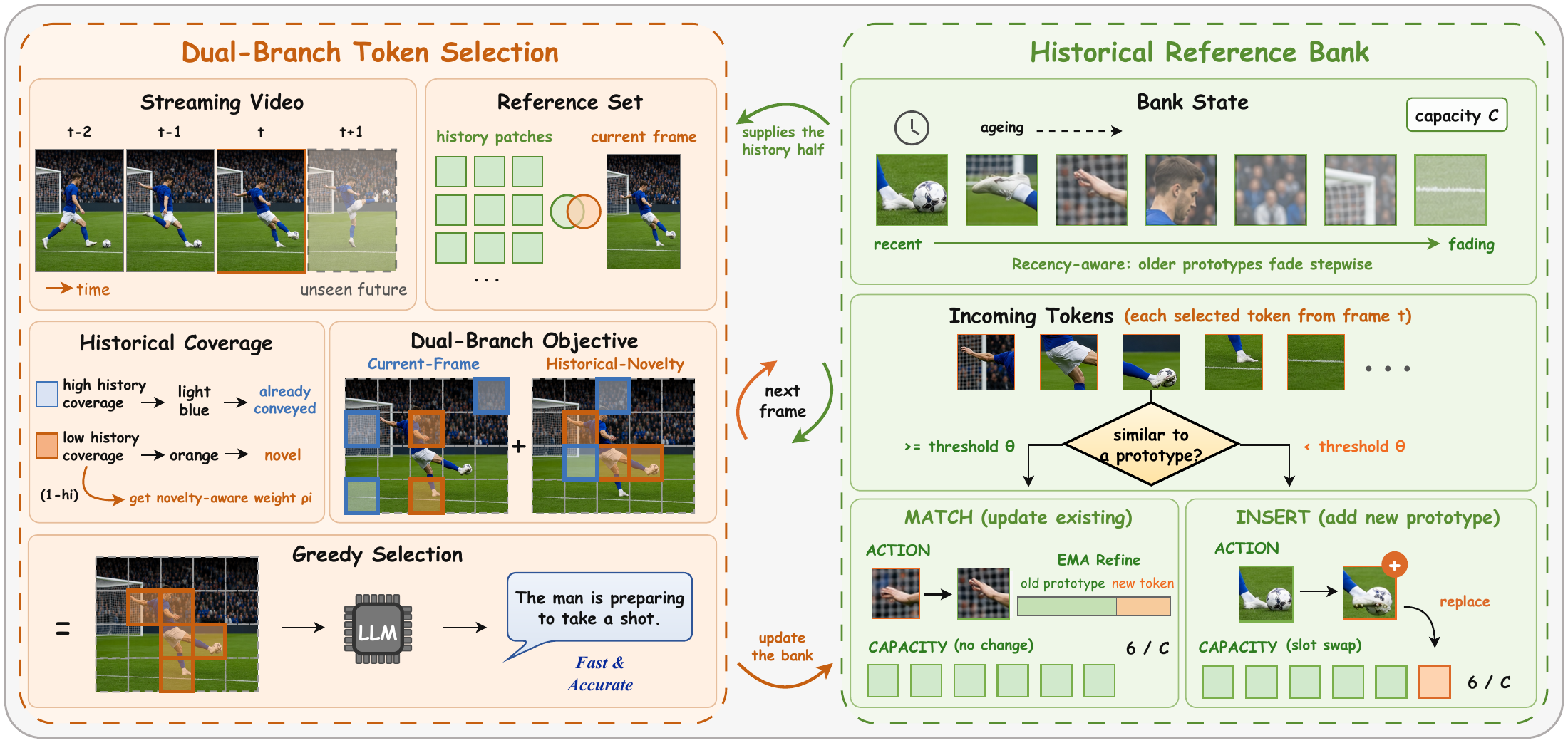}
\caption{Overview of NovaCov. A per-frame cycle between token selection
and bank update. Left: each current-frame token is
scored by its historical coverage $h_i$ (light blue: already conveyed; orange:
novel), which also yields the novelty-aware weight $\rho_i$; the dual-branch
objective is then greedily maximized to retain $K$ tokens. Right: those
tokens update the Bank, refining the nearest prototype by EMA when the similarity
reaches $\theta$ and inserting a new one otherwise, with a recency-aware utility
evicting the least useful prototype once capacity $C$ is exceeded.}
\label{fig:framework}
\end{figure*}

\section{Method}
 
\subsection{Preliminary and Problem Formulation}
 
\paragraph{Streaming inference pipeline.}
A VideoLLM is adapted to the streaming setting~\citep{di2025rekv} by processing the video incrementally, prefilling each arriving frame into the LLM. Formally, a video arrives as a stream $\mathcal{V}=\{v_t\}_{t=1}^{T}$, where each frame $v_t$ is encoded by a vision encoder and a projector~\citep{li2024llavaonevision} into $N$ visual tokens $\mathbf{X}_t=\{x_1,\dots,x_N\}$ with $x_i\in\mathbb{R}^{D}$, and we write $\hat{x}_i=x_i/\|x_i\|_2$ for the $\ell_2$-normalized token. Prefilling appends the key-value states of these tokens to a continually growing cache. In this setting, compression must be causal and query-agnostic: the decision at frame $t$ may depend only on $\mathbf{X}_t$ and on the history.
 
\paragraph{Problem formulation.}
Since a video stream has no known temporal boundary, the token sequence fed to the LLM grows with the stream. Streaming token compression therefore selects, at every arriving frame, an index subset $S_t\subseteq\{1,\dots,N\}$ of fixed size $|S_t|=K\ll N$ and forwards only the corresponding tokens $\{{x}_i\}_{i\in S_t}$, so that the sequence grows at a rate $K/N$ of the uncompressed one. Writing $F_t$ for an objective that scores a candidate subset at frame $t$, the problem is
\begin{equation}
S_t\;\in\;\argmax_{S\subseteq\{1,\dots,N\},\;|S|=K}F_t(S).
\label{eq:program}
\end{equation}
We construct $F_t$ from a single principle, coverage: the selected tokens should cover the current frame as well as the budget allows, while each token is credited only for what it adds to the content already accounted for, which we call the \emph{reference set}.
 
\paragraph{Facility-location coverage.}
We measure coverage with the facility-location function, a classical objective in subset selection~\citep{cornuejols1977location,lin2011class,krause2014submodular}. Given a ground set $V$ and a non-negative similarity matrix $A$, the coverage is defined as
\begin{equation}
f(S)=\sum_{i\in V}\max_{s\in S}A(i,s).
\label{eq:floc}
\end{equation}
Each element $i$ contributes its similarity to the most similar selected element, so a larger $f(S)$ means that $S$ represents $V$ better as a whole. Crucially, $f$ exhibits diminishing marginal gains: once an element is covered by $S$, adding a similar one yields little, which suppresses redundant selection by construction.
 
\paragraph{Reference set.}
Coverage involves two distinct sets, which must be kept apart. The set to be covered at frame $t$ is the current frame $\mathbf{X}_t$ alone, since tokens forwarded earlier are already held by the LLM and need not be represented again. The reference set is precisely that held content, against which a candidate earns credit only for what it adds, and in the streaming setting it comprises every token forwarded so far:
\begin{equation}
R_t^{\star}=\underbrace{\bigcup_{\tau<t}\{\hat{x}_i\}_{i\in S_\tau}}_{\text{history}}\;\cup\;\underbrace{\{\hat{x}_i\}_{i\in S_t}}_{\text{current selection}}.
\label{eq:ideal-ref}
\end{equation}
Its first term is fixed once the frame begins, whereas the second grows with every selection: a token just chosen joins the reference for those chosen after it, and when the frame is complete the whole of $S_t$ passes into the history term of frame $t+1$. By construction $R_t^{\star}$ is causally complete, yet its history term grows with the stream, so using it directly would make the per-frame cost rise without limit: this is the reference-set dilemma in concrete form. Our design therefore makes the reference set carry as much of the content of $R_t^{\star}$ as a bounded size allows.

Concretely, NovaCov runs a per-frame cycle (Figure~\ref{fig:framework}): the Bank carried over from $t-1$ supplies the history term, the dual-branch objective selects $K$ tokens from frame $t$, and those tokens are folded back into the Bank for $t+1$.
 
\subsection{Historical Reference Bank}
 
NovaCov summarizes the history term of $R_t^{\star}$ with a lightweight Historical Reference Bank $\mathcal{B}$ of prototypes $\{m_j\}$, so the reference set it actually uses at frame $t$ is $R_t=\{m_j\}_{j=1}^{|\mathcal{B}|}\cup\{\hat{x}_i\}_{i\in S_t}$, mirroring Eq.~(\ref{eq:ideal-ref}) term by term: the Bank is not the whole reference set but its bounded historical component. Serving compression alone and never entering the LLM, it records what past frames have retained, and its fixed capacity $C$ resolves the dilemma within a constant per-frame budget.
 
\paragraph{Bank representation.}
We represent $\mathcal{B}$ as a set of prototypes, $\mathcal{B}=\{(m_j,\tau_j,n_j)\}_{j=1}^{|\mathcal{B}|}$, with $|\mathcal{B}|\le C$. Here $m_j\in\mathbb{R}^{D}$ is a normalized prototype vector summarizing a cluster of historically similar retained tokens, $\tau_j$ records the frame index of its most recent update, and $n_j$ the cumulative number of times it has been matched. The Bank starts empty at $t=1$ and is populated online by the update below. We keep multiple prototypes so that a current token is tested against a \emph{specific} piece of past content.
 
\paragraph{Match-or-insert update.}
After the compression of each frame, the selected tokens are used to update the Bank online. For each selected token $\hat{x}_i$, $i\in S_t$, we find its nearest prototype $j^\star=\arg\max_j \hat{x}_i^{\top}m_j$. If the similarity exceeds a threshold $\theta$, the content is already represented by the Bank, and we refine the matched prototype toward it by an exponential moving average (EMA):
\begin{equation}
m_{j^\star}\leftarrow\mathrm{normalize}\big((1-\alpha)\,m_{j^\star}+\alpha\,\hat{x}_i\big),
\label{eq:ema}
\end{equation}
where $\alpha$ is the EMA coefficient; its timestamp and count are then refreshed by $\tau_{j^\star}\!\leftarrow\!t$ and $n_{j^\star}\!\leftarrow\!n_{j^\star}\!+\!1$. Otherwise the token carries content not yet in the Bank and is inserted as a new prototype $(\hat{x}_i,t,1)$. The EMA refinement lets recurring content, such as a static background, consolidate into one sharp prototype rather than many slots.
 
\paragraph{Recency-aware eviction.}
A bounded Bank must choose what to keep, and the criterion is usefulness ahead rather than fidelity to the past: a prototype earns its slot by being likely to recur, since only then can it mark incoming tokens as already conveyed. Whenever $|\mathcal{B}|>C$, we therefore evict the prototype of least utility, which balances recency and frequency:
\begin{equation}
u_j=\underbrace{2^{-(t-\tau_j)/\lambda}}_{\gamma_j}\cdot\log\big(1+n_j\big),
\label{eq:utility}
\end{equation}
where $\gamma_j\in(0,1]$ is a time-decay factor with half-life $\lambda$, and $t-\tau_j$ is the number of frames elapsed since the prototype was last updated. Both terms estimate how likely a prototype is to be matched again: recently updated content is probably still in view, while content matched many times has proved persistent. The Bank thus maintains an evolving active context rather than a permanent archive, spending a bounded capacity where it helps the next frame most.
 
\subsection{Dual-Branch Token Selection}
 
With the Bank supplying the history term of $R_t$, we now build $F_t$ from two branches, one for each term of the reference set: a current-frame branch over the frame itself and a historical-novelty residual branch over what the history has yet to cover.
 
\paragraph{Current coverage.}
We measure similarity by the cosine between token embeddings, clamped to discard negative values so that the matrix stays non-negative: $A(i,j)=\mathrm{clamp}(\hat{x}_i^{\top}\hat{x}_j,0,1)$. Instantiating Eq.~(\ref{eq:floc}) with $V=\{1,\dots,N\}$ gives $c_i(S)=\max_{s\in S}A(i,s)$, with $c_i(\emptyset)=0$, the degree to which token $i$ is covered by the current selection $S$.
 
\paragraph{Historical coverage.}
To incorporate the history term of $R_t$, we measure the degree to which each token of the current frame is covered by the Bank:
\begin{equation}
h_i=\max_{1\le j\le|\mathcal{B}|}\Big[\mathrm{clamp}(\hat{x}_i^{\top}m_j,0,1)\cdot\gamma_j\Big]\in[0,1],
\label{eq:hist-cov}
\end{equation}
where $\gamma_j$ is the time-decay factor of Eq.~(\ref{eq:utility}). A value $h_i\to 1$ means token $i$ is highly similar to recently seen content and is already well covered by the history, whereas $h_i\to 0$ means it carries content absent from that history. When the Bank is empty, at the first frame, all $h_i=0$ and the objective reduces to intra-frame coverage alone.
 
\paragraph{Novelty-aware weighting.}
Within the current-frame branch, whose gains are not already discounted by the history, a novel token should command a higher coverage demand than a redundant one. We therefore weight each token by its temporal novelty $1-h_i$, rank-normalized within the frame for stability:
\begin{equation}
\rho_i=\rho_{\min}+(1-\rho_{\min})\cdot\mathrm{RankNorm}(1-h_i)\in[\rho_{\min},1],
\label{eq:rho}
\end{equation}
where $\mathrm{RankNorm}$ maps $\{1-h_i\}_{i=1}^{N}$ to normalized ranks in $[0,1]$, and the floor $\rho_{\min}>0$ keeps even a fully redundant token in demand, so that the spatial skeleton of the frame does not collapse.
 
\paragraph{Dual-Branch coverage objective.}
We regard $h_i$ as a \textbf{lower bound} on the coverage of token $i$: the current-frame selection contributes to the residual branch only when it lifts the coverage of $i$ above the historical level. Combining this branch with the novelty-weighted current-frame branch yields the objective of Eq.~(\ref{eq:program}),
\begin{equation}
F_t(S)=\lambda_{\mathrm{c}}\sum_{i=1}^{N}\rho_i\,c_i(S)
+\lambda_{\mathrm{r}}\sum_{i=1}^{N}\max\big(c_i(S)-h_i,\,0\big),
\label{eq:objective}
\end{equation}
where $\lambda_{\mathrm{c}},\lambda_{\mathrm{r}}\ge0$ balance the two branches. The two play complementary roles: the residual branch yields almost no gain for tokens already well covered ($h_i$ close to $1$) and thus steers the budget toward new content, while the current-frame branch, unconstrained by the historical lower bound, always rewards covering the current frame and so preserves its spatial structure.
 
\paragraph{Greedy selection.}
Maximizing a coverage objective under a cardinality constraint is NP-hard~\citep{feige1998threshold}, so we solve Eq.~(\ref{eq:program}) with the greedy algorithm: starting from $S=\emptyset$, each round adds the token of largest marginal gain until $|S|=K$, maintaining one running coverage state per branch, $c_i$ for the current-frame branch and $r_i$ for the residual one, initialized to $0$ and $h_i$ respectively. The selected tokens are output in the original raster order to keep the positions fed to the LLM consistent. Since $F_t$ is monotone submodular, this single pass is provably near-optimal.

\begin{table*}[t]
\centering
\small
\setlength{\tabcolsep}{1mm}
\begin{tabular*}{\textwidth}{@{\extracolsep{\fill}} l cccccccccc c c @{}}
\toprule
\textbf{Method} & CS & OP & ATP & PR & ACP & SU & EU & CT & TR & CR & \textbf{Overall} & \textbf{LLM Pref. Latency} \\
\midrule
ReKV$^{\dagger}$ (ICLR25) & 79.2 & 77.5 & 75.6 & 66.0 & 62.2 & 60.3 & 72.3 & 43.6 & 69.7 & 79.5
& 69.1 & 482.4 \\
\midrule
+\,ToMe (ICLR23) & 67.5 & 64.6 & 66.3 & 63.0 & 58.4 & 53.3 & \underline{65.2} & 19.7 & 57.3 & \underline{76.6}
& 59.4 (86.0\%) & 257.8 ($\downarrow$46.6\%) \\
+\,VisionZip (CVPR25) & 69.5 & 66.4 & 69.3 & 50.7 & 52.9 & \underline{57.2} & 64.1 & 33.3 & 54.8 & 73.4
& 60.4 (87.4\%) & 258.3 ($\downarrow$46.5\%) \\
+\,VidCom$^2$ (EMNLP25) & \underline{76.0} & \underline{68.1} & \underline{71.6} & 62.0 & 58.6 & 52.0 & 64.0 & 42.5 & 60.4 & \underline{76.6}
& 63.6 (92.0\%) & 259.1 ($\downarrow$46.3\%) \\
+\,STC-Pruner (CVPR26) & 75.4 & 66.8 & 71.2 & \underline{63.9} & \underline{57.8} & 51.2 & 64.0 & \textbf{45.1} & \underline{63.2} & \underline{76.6}
& \underline{63.7 (92.2\%)} & 259.2 ($\downarrow$46.3\%) \\
\textbf{+\,NovaCov (Ours)} & \textbf{75.7} & \textbf{74.5} & \textbf{77.6} & \textbf{71.3} & \textbf{62.5} & \textbf{58.5} & \textbf{68.0} & \underline{44.2} & \textbf{67.6} & \textbf{80.5}
& \textbf{68.4 (99.0\%)} & 259.9 ($\downarrow$46.2\%) \\
\bottomrule
\end{tabular*}
\caption{Comprehensive evaluation results on StreamingBench across real-time understanding tasks: $^{\dagger}$: the uncompressed upper bound. CS: Clips Summarization, OP: Object Perception, ATP: Attribute Perception, PR: Prospective Reasoning, ACP: Action Perception, SU: Spatial Understanding, EU: Event Understanding, CT: Counting, TR: Text-Rich Understanding, CR: Causal Reasoning. “LLM Pref. Latency” includes token compression and LLM prefilling (ms).}
\label{tab:streaming}
\end{table*}

\begin{table*}[t]
\centering
\small
\setlength{\tabcolsep}{0.8mm}
\begin{tabular*}{\textwidth}{@{\extracolsep{\fill}} l ccccccc cccc cccc c c @{}}
\toprule
 & \multicolumn{7}{c}{\shortstack{\textbf{Real-Time Visual}\\\textbf{Perception}}}
 & \multicolumn{4}{c}{\shortstack{\textbf{Backward}\\\textbf{Tracing}}}
 & \multicolumn{4}{c}{\shortstack{\textbf{Forward Active}\\\textbf{Responding}}}
 & & \textbf{LLM Pref.} \\
\cmidrule(lr){2-8}\cmidrule(lr){9-12}\cmidrule(lr){13-16}
\textbf{Method} & OCR & ACR & ATR & STU & FPD & OJR & Avg.
 & EPM & ASI & HLD & Avg. & REC & SSR & CRR & Avg. & \textbf{Overall} & \textbf{Latency} \\
\midrule
ReKV$^{\dagger}$ (ICLR25) & 73.8 & 56.0 & 74.1 & 51.7 & 70.3 & 60.3 & 64.4 & 54.2 & 57.4 & 28.5 & 46.7
& 25.4 & 64.6 & 53.3 & 47.8
& 52.9 & 482.4 \\
\midrule
+\,ToMe (ICLR23) & 61.1 & 49.5 & 52.6 & 42.7 & 62.4 & 50.0 & 53.1
& 49.2 & 50.0 & 29.6 & 42.9
& 19.2 & 60.7 & 50.0 & 43.3
& 46.4  & 257.8 \\
+\,VisionZip (CVPR25) & 49.0 & 49.5 & 64.7 & 44.4 & 64.4 & 50.5 & 53.8 & 47.1 & \underline{54.1} & 30.7 & 44.0
& 21.9 & 58.4 & \textbf{53.8} & 44.7
& 47.5 & 258.3 \\
+\,VidCom$^2$ (EMNLP25) & \underline{65.8} & \textbf{59.6} & \underline{69.0} & 47.2 & 64.4 & \underline{56.5} & 60.4
& 50.5 & 53.4 & \underline{32.8} & \underline{45.6}
& 25.8 & 59.0 & 50.8 & 45.2
& 50.4 & 259.1 \\
+\,STC-Pruner (CVPR26) & 64.4 & \textbf{59.6} & 68.1 & \textbf{48.9} & \underline{65.4} & \underline{56.5} & \underline{60.5}
& \underline{51.2} & 52.0 & \textbf{33.3} & 45.5
& \underline{25.9} & \underline{59.3} & \underline{52.1} & \underline{45.8}
& \underline{50.6} & 259.2 \\
\textbf{+\,NovaCov (Ours)} & \textbf{67.1} & \underline{57.2} & \textbf{69.8} & \underline{48.5} & \textbf{73.3} & \textbf{59.8} & \textbf{62.4}
& \textbf{52.5} & \textbf{56.1} & 31.7 & \textbf{46.8}
& \textbf{28.8} & \textbf{64.9} & \underline{52.1} & \textbf{48.6}
& \textbf{52.7} & 259.9 \\

\bottomrule
\end{tabular*}
\caption{Comprehensive evaluation results on OVO-Bench across three categories: (i) Real-Time Visual Perception (OCR: Optical Character Recognition, ACR: Action Recognition, ATR: Attribute Recognition, STU: Spatial Understanding, FPD: Future Prediction, OJR: Object Recognition), (ii) Backward Tracing (EPM: Episodic Memory, ASI: Action Sequence Identification, HLD: Hallucination Detection), (iii) Forward Active Responding (REC: Repetition Event Count, SSR: Sequential Steps Recognition, CRR: Clues Reveal Responding).}
\label{tab:ovo}
\end{table*}
 
\subsection{Theoretical Analysis}
 
We next prove that enriching the objective with a history term and novelty-aware weights preserves the classical greedy approximation guarantee.
 
\paragraph{Definitions.}
For $f:2^{V}\to\mathbb{R}$ and all $S\subseteq T\subseteq V$, $a\in V\setminus T$:
$f$ is \emph{normalized} if $f(\emptyset)=0$; \emph{monotone} if $f(S)\le f(T)$;
and \emph{submodular} if $f(S\cup\{a\})-f(S)\ \ge\ f(T\cup\{a\})-f(T)$, marginal
gains never growing as the selection grows. We use the classical fact that the
weighted facility-location function
$g_{W,w}(S)=\sum_{i\in V}w_i\max_{s\in S}W(i,s)$ is normalized, non-negative,
monotone and submodular for non-negative $W$ and
$w$~\citep{krause2014submodular}.
 
\paragraph{Proposition 1.}
\emph{For $\lambda_{\mathrm{c}},\lambda_{\mathrm{r}}\ge0$, the objective $F_t$ in
Eq.~(\ref{eq:objective}) is normalized, non-negative, monotone and submodular on
$2^{V}$, and greedy selection therefore returns $\hat{S}$ with
$F_t(\hat{S})\ge(1-1/e)\max_{|S|\le K}F_t(S)$.}
 
\noindent\textit{Proof.}
At frame $t$ the quantities $A$, $h$ and $\rho$ are fixed before selection begins
and are therefore constant in $S$. The current-frame branch is $g_{A,\rho}$, as
$A\ge0$ and $\rho_i\ge\rho_{\min}>0$. Since $z\mapsto\max(z-h_i,0)$ is monotone,
it commutes with the maximum over a finite non-empty $S$, so
$\max(c_i(S)-h_i,0)=\max_{s\in S}\tilde{A}_h(i,s)$ with
$\tilde{A}_h(i,s)=\max(A(i,s)-h_i,0)\ge0$, both sides vanishing for
$S=\emptyset$; the residual branch is thus $g_{\tilde{A}_h,\mathbf{1}}$. A
non-negative combination preserves every property, and the bound then follows
from the classical guarantee for monotone submodular maximization under a
cardinality constraint~\citep{nemhauser1978analysis}. \hfill$\square$
 
The two branches are thus one function class on two matrices: $A$, and
$\tilde{A}_h$ thresholded by the historical coverage. History enters as a shift of
the similarity rather than an external penalty, so the richer reference set costs
nothing in approximation quality, and no polynomial-time algorithm can do better
unless $\mathrm{P}=\mathrm{NP}$~\citep{feige1998threshold}, a guarantee that
independent token ranking cannot claim.

\section{Experiments}
\label{sec:experiments}

\subsection{Experimental Setup}
 
\paragraph{Benchmarks.}
We evaluate NovaCov across two benchmark categories: \textbf{(i)}~\emph{Streaming video understanding}: we use OVO-Bench~\citep{niu2025ovobench} and StreamingBench~\citep{lin2026streamingbench} to assess performance in streaming scenarios. \textbf{(ii)}~\emph{Long video understanding}: we select EgoSchema~\citep{mangalam2023egoschema}, MLVU-dev~\citep{zhou2025mlvu}, and VideoMME~\citep{fu2025videomme} to evaluate effectiveness in offline long-video understanding.
 
\paragraph{Implementation details.}
We adopt ReKV~\citep{di2025rekv} with LLaVA-OV-7B~\citep{li2024llavaonevision} as the backbone, following its original settings (0.5\,fps protocol). All compression methods are plugged into the same backbone, replacing only the pre-LLM token selection and leaving the rest of the pipeline unchanged: of the $N=196$ visual tokens per frame, $K=50$ are retained, compressing the sequence to about 25\%. For NovaCov we set the Reference Bank capacity $C=512$, half-life $\lambda=16$ frames, match threshold $\theta=0.9$, EMA coefficient $\alpha=0.10$, branch weights $\lambda_{\mathrm{c}}=0.1$ and $\lambda_{\mathrm{r}}=0.9$, and novelty floor $\rho_{\min}=0.20$. Both branches are fused into a single Triton kernel~\citep{tillet2019triton}, so each frame is selected in one pass. All experiments run on a single 32GB NVIDIA GeForce RTX 5090 GPU.
 
\paragraph{Baselines.}
We compare NovaCov against four training-free token compression methods: ToMe~\citep{bolya2023tome}, VisionZip~\citep{yang2025visionzip}, VidCom$^2$~\citep{liu2025video}, and STC-Pruner~\citep{wang2025stc}, all of which score tokens in a token-wise manner and among which STC-Pruner is the strongest streaming pruner to date. Existing set-wise methods assume a fully visible video and cannot be transferred directly to streaming, so we instead examine the set-wise setting through the Current-only ablation in Table~\ref{tab:ablation}.

\begin{table*}[t]
\centering
\small
\setlength{\tabcolsep}{1mm}
\begin{tabular*}{\textwidth}{@{\extracolsep{\fill}} l cc cccc c @{}}
\toprule
 & & & \multicolumn{4}{c}{\textbf{VideoMME}} & \\
\cmidrule(lr){4-7}
\textbf{Method} & \textbf{EgoSchema} & \textbf{MLVU-dev} & Short & Med. & Long & Overall & \textbf{Avg.} \\
\midrule
ReKV$^{\dagger}$ (ICLR25) & 57.7 & 68.6 & 70.4 & 55.4 & 47.3 & 57.7 & 61.3 \\
\midrule
+\,ToMe (ICLR23) & 55.2 & 63.1 & 59.6 & 52.0 & 43.4 & 51.7 & 56.7 \\
+\,VisionZip (CVPR25) & 55.8 & 63.2 & 59.6 & 51.8 & 43.4 & 51.6 & 56.9 \\
+\,VidCom$^2$ (EMNLP25) & 60.6 & 67.1 & \underline{68.2} & 55.7 & \underline{46.6} & 56.8 & 61.5 \\
+\,STC-Pruner (CVPR26) & \underline{60.8} & \underline{67.6} & \textbf{68.7} & \underline{56.3} & 46.3 & \underline{57.1} & \underline{61.8} \\
\textbf{+\,NovaCov (Ours)} & \textbf{61.4} & \textbf{67.9} & \textbf{68.7} & \textbf{57.3} & \textbf{46.6} & \textbf{57.5} & \textbf{62.3} \\
\bottomrule
\end{tabular*}
\caption{Quantitative evaluation results on three offline long video understanding benchmarks.}
\label{tab:offline}
\end{table*}

\subsection{Main Comparisons}
 
Tables~\ref{tab:streaming},~\ref{tab:ovo},~\ref{tab:offline} present comprehensive comparisons of NovaCov against other compression methods in streaming and long video understanding scenarios, addressing two questions: \textbf{(i)}~\emph{how it performs across diverse scenarios?} \textbf{(ii)}~\emph{how the run-time cost of its set-wise objective trades off against the accuracy it delivers?}
 
\paragraph{Performance across diverse scenarios.}
NovaCov outperforms every compression method on all five benchmarks. Against the previous state of the art STC-Pruner~\citep{wang2025stc}, it improves by 4.7 on StreamingBench and 2.1 on OVO-Bench, retaining 99.0\% and 99.6\% of the uncompressed ReKV, and it leads on all three offline benchmarks for an average of 62.3 against 61.8. These results showcase the strong performance of NovaCov in streaming scenarios and demonstrate its robust performance across diverse settings. The reference set still matters offline, only less visibly: a complete video answers the question trivially rather than making it irrelevant.
 
More striking is where the budget lands. On five of the ten StreamingBench tasks NovaCov exceeds the uncompressed upper bound, and the excess concentrates on cross-frame reasoning: Prospective Reasoning reaches 71.3 against ReKV's 66.0, Attribute Perception 77.6 against 75.6, and Repetition Event Count on OVO-Bench 28.8 against 25.4. Feeding the model every token is evidently not the same as feeding it every useful one; once redundant background is gone, what remains competes for attention against far less noise.
 
\paragraph{Accuracy--latency trade-off.}
At a fixed budget the latency question is nearly degenerate: all compression methods prefill within 2.1\,ms of one another, a spread of 0.8\%, while their accuracies differ by up to 9.0 points. Greedy maximization needs only $\mathcal{O}(NK)$ marginal-gain evaluations~\citep{nemhauser1978analysis,cho2026floc}, and fusing both branches into a single Triton kernel~\citep{tillet2019triton} brings the measured cost to the level of token-wise baselines: 259.9\,ms against 482.4\,ms uncompressed, a reduction of 46.2\%. The set-wise objective therefore costs no measurable latency while returning 4.7 points more than the next best method.

\begin{table}[t]
\centering
\small
\setlength{\tabcolsep}{0.6mm}
\begin{tabular*}{\columnwidth}{
    @{\extracolsep{\fill}} l ccc ccc c @{}
}
\toprule
& \multicolumn{3}{c}{\textbf{Component}}
& \multicolumn{3}{c}{\textbf{OVO-Bench}}
& \multicolumn{1}{c}{
    \textbf{Streaming}}
 \\
\cmidrule(lr){2-4}
\cmidrule(lr){5-7}
& Bank & Hist. & Nov.
& EPM & FPD & REC
& \multicolumn{1}{c}{
    \textbf{Bench}}
 \\
\midrule
(a) Current-only
& $\times$ & $\times$ & $\times$
& 52.1 & 68.6 & 27.1
& 64.2 \\

(b) + Bank
& $\checkmark$ & $\checkmark$ & $\times$
& 52.4 & 72.4 & 27.9
& 67.7 \\

(c) + Novelty
& $\checkmark$ & $\checkmark$ & $\checkmark$
& \textbf{52.5} & \textbf{73.3} & \textbf{28.8}
& \textbf{68.4} \\

\midrule
(d) Unbounded
& $\infty$ & $\checkmark$ & $\checkmark$
& 51.7 & 70.2 & 27.7
& 67.4 \\
\bottomrule
\end{tabular*}

\caption{
Ablation on three OVO-Bench subsets and StreamingBench overall ("Bank": bounded Historical Reference Bank; "Hist.": historical-novelty branch;
"Nov.": novelty-aware weighting). Variant~(a) uses the current coverage branch alone; (c) is the full NovaCov; (d) never evicts, so the Bank grows without
bound.
}
\label{tab:ablation}
\end{table}

\subsection{Ablation Study}
\label{sec:ablation}
 
We ablate NovaCov on StreamingBench and three representative OVO-Bench subsets—FPD, EPM, and REC—each representing one of its three major task categories, using the same 25\% retention ratio as in the main comparisons.
 
\paragraph{Reference set.}
Even with the current frame as its only reference~(a), the set-wise objective already reaches 64.2 on StreamingBench, above the 63.7 of the state-of-the-art token-wise pruner. The decisive step, however, is the reference set itself. Adding the bounded Bank~(b) supplies the history term and raises StreamingBench to 67.7, with the largest subset gain on the Future Prediction task (68.6 to 72.4), where anticipating what comes next depends on what earlier frames established. What a set-wise method is able to reference, rather than the formulation itself, is what carries the improvement.
 
\paragraph{Novelty weighting.}
Weighting current-frame demand by temporal novelty~(c) raises StreamingBench further to 68.4, with Repetition Event Count rising from 27.9 to 28.8. Counting repeated events depends on registering each new occurrence, which is exactly what a novelty-weighted budget preserves.
 
\paragraph{Bounded vs.\ unbounded.}
Variant~(d) lifts the capacity limit, so the Bank never evicts and grows with the stream. Its cost rises without bound, and accuracy falls as well, from 68.4 to 67.4 on StreamingBench: a reference that forgets nothing finds a near-match for almost any incoming token, so historical coverage saturates and content that faded and returned is no longer recognized as new. The bounded update therefore meets the real-time requirement at no cost in accuracy: what a reference set needs is alignment with the active context, not exhaustive recall.

\section{Conclusion}
 
In this work, we introduce NovaCov, a training-free, plug-and-play component and, to our knowledge, the first set-wise token compressor for streaming video. NovaCov pairs a capacity-bounded, recency-weighted Historical Reference Bank with a dual-branch coverage objective over a current-frame branch and a historical-novelty branch. Both rest on one idea: a token is worth keeping for what it adds to the content the model already holds, which makes the reference set the crux and, under streaming, requires it to stay bounded while still tracking a growing context. As both branches are facility-location functions, greedy selection retains the classical $(1-1/e)$ guarantee. NovaCov attains state-of-the-art accuracy on streaming and offline benchmarks at latency comparable to strong baselines.

\bibliography{references}

\begin{thebibliography}{41}
\providecommand{\natexlab}[1]{#1}

\bibitem[{Bai et~al.(2025)Bai, Chen, Liu, Wang, Ge, Song, Dang, Wang, Wang, Tang, Zhong, Zhu, Yang, Li, Wan, Wang, Ding, Fu, Xu, Ye, Zhang, Xie, Cheng, Zhang, Yang, Xu, and Lin}]{bai2025qwen25vl}
Bai, S.; Chen, K.; Liu, X.; Wang, J.; Ge, W.; Song, S.; Dang, K.; Wang, P.; Wang, S.; Tang, J.; Zhong, H.; Zhu, Y.; Yang, M.; Li, Z.; Wan, J.; Wang, P.; Ding, W.; Fu, Z.; Xu, Y.; Ye, J.; Zhang, X.; Xie, T.; Cheng, Z.; Zhang, H.; Yang, Z.; Xu, H.; and Lin, J. 2025.
\newblock {Qwen2.5-VL} Technical Report.
\newblock arXiv:2502.13923.

\bibitem[{Bolya et~al.(2023)Bolya, Fu, Dai, Zhang, Feichtenhofer, and Hoffman}]{bolya2023tome}
Bolya, D.; Fu, C.-Y.; Dai, X.; Zhang, P.; Feichtenhofer, C.; and Hoffman, J. 2023.
\newblock Token Merging: Your {ViT} but Faster.
\newblock In \emph{International Conference on Learning Representations}.

\bibitem[{Chen et~al.(2024)Chen, Zhao, Liu, Bai, Lin, Zhou, and Chang}]{chen2024fastv}
Chen, L.; Zhao, H.; Liu, T.; Bai, S.; Lin, J.; Zhou, C.; and Chang, B. 2024.
\newblock An Image is Worth 1/2 Tokens After Layer 2: Plug-and-Play Inference Acceleration for Large Vision-Language Models.
\newblock In \emph{European Conference on Computer Vision (ECCV)}, 19--35.

\bibitem[{Chen et~al.(2025{\natexlab{a}})Chen, Tao, Shao, and Wang}]{chen2025streamingtom}
Chen, X.; Tao, K.; Shao, K.; and Wang, H. 2025{\natexlab{a}}.
\newblock Streamingtom: Streaming token compression for efficient video understanding.
\newblock \emph{arXiv preprint arXiv:2510.18269}.

\bibitem[{Chen et~al.(2025{\natexlab{b}})Chen, Xue, Li, Hu, Zhu, Li, Fang, Tang, Yang, Liu, He, Yin, Molchanov, Kautz, Fan, Zhu, Lu, and Han}]{chen2025longvila}
Chen, Y.; Xue, F.; Li, D.; Hu, Q.; Zhu, L.; Li, X.; Fang, Y.; Tang, H.; Yang, S.; Liu, Z.; He, E.; Yin, H.; Molchanov, P.; Kautz, J.; Fan, J.; Zhu, Y.; Lu, Y.; and Han, S. 2025{\natexlab{b}}.
\newblock {LongVILA}: Scaling Long-Context Visual Language Models for Long Videos.
\newblock In \emph{International Conference on Learning Representations}.

\bibitem[{Cho et~al.(2026)Cho, Lee, Hayat, Hwang, Porikli, and Choi}]{cho2026floc}
Cho, J.; Lee, J.; Hayat, M.; Hwang, K.; Porikli, F.; and Choi, S. 2026.
\newblock FLoC: Facility Location-Based Efficient Visual Token Compression for Long Video Understanding.
\newblock In \emph{International Conference on Learning Representations (ICLR)}.

\bibitem[{Cornu{\'e}jols, Fisher, and Nemhauser(1977)}]{cornuejols1977location}
Cornu{\'e}jols, G.; Fisher, M.~L.; and Nemhauser, G.~L. 1977.
\newblock Location of Bank Accounts to Optimize Float: An Analytic Study of Exact and Approximate Algorithms.
\newblock \emph{Management Science}, 23(8): 789--810.

\bibitem[{Di et~al.(2025)Di, Yu, Zhang, Li, Zhong, Cheng, Li, He, Shu, and Jiang}]{di2025rekv}
Di, S.; Yu, Z.; Zhang, G.; Li, H.; Zhong, T.; Cheng, H.; Li, B.; He, W.; Shu, F.; and Jiang, H. 2025.
\newblock Streaming Video Question-Answering with In-Context Video {KV}-Cache Retrieval.
\newblock In \emph{International Conference on Learning Representations}.

\bibitem[{Dong et~al.(2025)Dong, Hu, Zhang, Yin, Fu, and Qian}]{dong2025mmtok}
Dong, S.; Hu, J.; Zhang, M.; Yin, M.; Fu, Y.; and Qian, Q. 2025.
\newblock Mmtok: Multimodal coverage maximization for efficient inference of vlms.
\newblock \emph{arXiv preprint arXiv:2508.18264}.

\bibitem[{Feige(1998)}]{feige1998threshold}
Feige, U. 1998.
\newblock A Threshold of $\ln n$ for Approximating Set Cover.
\newblock \emph{Journal of the ACM}, 45(4): 634--652.

\bibitem[{Fu et~al.(2025)Fu, Dai, Luo, Li, Ren, Zhang, Wang, Zhou, Shen, Zhang, Chen, Li, Lin, Zhao, Li, Xu, Zheng, Chen, Shan, He, and Sun}]{fu2025videomme}
Fu, C.; Dai, Y.; Luo, Y.; Li, L.; Ren, S.; Zhang, R.; Wang, Z.; Zhou, C.; Shen, Y.; Zhang, M.; Chen, P.; Li, Y.; Lin, S.; Zhao, S.; Li, K.; Xu, T.; Zheng, X.; Chen, E.; Shan, C.; He, R.; and Sun, X. 2025.
\newblock {Video-MME}: The First-Ever Comprehensive Evaluation Benchmark of Multi-Modal {LLM}s in Video Analysis.
\newblock In \emph{Proceedings of the IEEE/CVF Conference on Computer Vision and Pattern Recognition}, 24108--24118.

\bibitem[{Kim et~al.(2026)Kim, Shim, Choi, and Chang}]{kim2026infinipot}
Kim, M.; Shim, K.; Choi, J.; and Chang, S. 2026.
\newblock Infinipot-v: Memory-constrained kv cache compression for streaming video understanding.
\newblock \emph{Advances in Neural Information Processing Systems}, 38: 138983--139013.

\bibitem[{Krause and Golovin(2014)}]{krause2014submodular}
Krause, A.; and Golovin, D. 2014.
\newblock Submodular function maximization.
\newblock \emph{Tractability}, 3(71-104): 3.

\bibitem[{Lee, Wen, and Choi(2026)}]{lee2026moving}
Lee, J.; Wen, S.; and Choi, D.-W. 2026.
\newblock Moving Beyond Diversity: Visual Token Pruning as Subspace Reconstruction for Efficient VLMs.
\newblock \emph{arXiv preprint arXiv:2606.18681}.

\bibitem[{Li et~al.(2024{\natexlab{a}})Li, Zhang, Guo, Zhang, Li, Zhang, Zhang, Zhang, Li, Liu, and Li}]{li2024llavaonevision}
Li, B.; Zhang, Y.; Guo, D.; Zhang, R.; Li, F.; Zhang, H.; Zhang, K.; Zhang, P.; Li, Y.; Liu, Z.; and Li, C. 2024{\natexlab{a}}.
\newblock {LLaVA-OneVision}: Easy Visual Task Transfer.
\newblock arXiv:2408.03326.

\bibitem[{Li et~al.(2024{\natexlab{b}})Li, Wang, He, Li, Wang, Liu, Wang, Xu, Chen, Luo, Wang, and Qiao}]{li2024mvbench}
Li, K.; Wang, Y.; He, Y.; Li, Y.; Wang, Y.; Liu, Y.; Wang, Z.; Xu, J.; Chen, G.; Luo, P.; Wang, L.; and Qiao, Y. 2024{\natexlab{b}}.
\newblock {MVBench}: A Comprehensive Multi-Modal Video Understanding Benchmark.
\newblock In \emph{Proceedings of the IEEE/CVF Conference on Computer Vision and Pattern Recognition}, 22195--22206.

\bibitem[{Lin and Bilmes(2011)}]{lin2011class}
Lin, H.; and Bilmes, J. 2011.
\newblock A Class of Submodular Functions for Document Summarization.
\newblock In \emph{Proceedings of the 49th Annual Meeting of the Association for Computational Linguistics: Human Language Technologies}, 510--520.

\bibitem[{Lin et~al.(2026)Lin, Fang, Chen, Cheng, Wan, Luo, Wang, Li, Liu, and Sun}]{lin2026streamingbench}
Lin, J.; Fang, Z.; Chen, C.; Cheng, H.; Wan, Z.; Luo, F.; Wang, Z.; Li, P.; Liu, Y.; and Sun, M. 2026.
\newblock Streamingbench: Assessing the gap for mllms to achieve streaming video understanding.
\newblock In \emph{ICASSP 2026-2026 IEEE International Conference on Acoustics, Speech and Signal Processing (ICASSP)}, 12147--12151. IEEE.

\bibitem[{Liu et~al.(2024)Liu, Yu, Lan, Wang, Fang, Kautz, Li, and Alvarez}]{liu2024streamchat}
Liu, J.; Yu, Z.; Lan, S.; Wang, S.; Fang, R.; Kautz, J.; Li, H.; and Alvarez, J.~M. 2024.
\newblock {StreamChat}: Chatting with Streaming Video.
\newblock arXiv:2412.08646.

\bibitem[{Liu et~al.(2025)Liu, Wang, Ma, and Zhang}]{liu2025video}
Liu, X.; Wang, Y.; Ma, J.; and Zhang, L. 2025.
\newblock Video compression commander: Plug-and-play inference acceleration for video large language models.
\newblock In \emph{Proceedings of the 2025 Conference on Empirical Methods in Natural Language Processing}, 1910--1924.

\bibitem[{Mangalam, Akshulakov, and Malik(2023)}]{mangalam2023egoschema}
Mangalam, K.; Akshulakov, R.; and Malik, J. 2023.
\newblock Egoschema: A diagnostic benchmark for very long-form video language understanding.
\newblock \emph{Advances in Neural Information Processing Systems}, 36: 46212--46244.

\bibitem[{Nemhauser, Wolsey, and Fisher(1978)}]{nemhauser1978analysis}
Nemhauser, G.~L.; Wolsey, L.~A.; and Fisher, M.~L. 1978.
\newblock An Analysis of Approximations for Maximizing Submodular Set Functions---{I}.
\newblock \emph{Mathematical Programming}, 14(1): 265--294.

\bibitem[{Ning et~al.(2025)Ning, Liu, Jin, Ding, Guo, and Zhao}]{ning2025livevlm}
Ning, Z.; Liu, G.; Jin, Q.; Ding, W.; Guo, M.; and Zhao, J. 2025.
\newblock {LiveVLM}: Efficient Online Video Understanding via Streaming-Oriented {KV} Cache and Retrieval.
\newblock arXiv:2505.15269.

\bibitem[{Niu et~al.(2025)Niu, Li, Miao, Ge, Zhou, He, Dong, Duan, Ding, Qian, Zhang, Zang, Cao, He, and Wang}]{niu2025ovobench}
Niu, J.; Li, Y.; Miao, Z.; Ge, C.; Zhou, Y.; He, Q.; Dong, X.; Duan, H.; Ding, S.; Qian, R.; Zhang, P.; Zang, Y.; Cao, Y.; He, C.; and Wang, J. 2025.
\newblock {OVO-Bench}: How Far Are Your {Video-LLM}s from Real-World Online Video Understanding?
\newblock In \emph{Proceedings of the IEEE/CVF Conference on Computer Vision and Pattern Recognition}, 18902--18913.

\bibitem[{Ren et~al.(2023)Ren, Chen, Li, Sun, and Hou}]{ren2023testa}
Ren, S.; Chen, S.; Li, S.; Sun, X.; and Hou, L. 2023.
\newblock TESTA: Temporal-Spatial Token Aggregation for Long-form Video-Language Understanding.
\newblock In \emph{Findings of the Association for Computational Linguistics: EMNLP 2023}, 932--947.

\bibitem[{Shao et~al.(2025)Shao, Tao, Qin, You, Sui, and Wang}]{shao2025holitom}
Shao, K.; Tao, K.; Qin, C.; You, H.; Sui, Y.; and Wang, H. 2025.
\newblock HoliTom: Holistic Token Merging for Fast Video Large Language Models.
\newblock In \emph{Advances in Neural Information Processing Systems (NeurIPS)}.

\bibitem[{Shen et~al.(2025)Shen, Gong, He, Zhang, Liu, Zhao, and Ding}]{shen2025fastvid}
Shen, L.; Gong, G.; He, T.; Zhang, Y.; Liu, P.; Zhao, S.; and Ding, G. 2025.
\newblock FastVID: Dynamic Density Pruning for Fast Video Large Language Models.
\newblock In \emph{Advances in Neural Information Processing Systems (NeurIPS)}.

\bibitem[{Tao et~al.(2025)Tao, Qin, You, Sui, and Wang}]{tao2025dycoke}
Tao, K.; Qin, C.; You, H.; Sui, Y.; and Wang, H. 2025.
\newblock DyCoke: Dynamic Compression of Tokens for Fast Video Large Language Models.
\newblock In \emph{Proceedings of the IEEE/CVF Conference on Computer Vision and Pattern Recognition (CVPR)}.

\bibitem[{Tillet, Kung, and Cox(2019)}]{tillet2019triton}
Tillet, P.; Kung, H.~T.; and Cox, D. 2019.
\newblock Triton: An Intermediate Language and Compiler for Tiled Neural Network Computations.
\newblock In \emph{Proceedings of the 3rd ACM SIGPLAN International Workshop on Machine Learning and Programming Languages (MAPL)}, 10--19.

\bibitem[{Wang et~al.(2026{\natexlab{a}})Wang, Chen, Huang, Li, Li, Liu, Kautz, Alvarez, Zhang, and Yu}]{wang2026videoitg}
Wang, S.; Chen, G.; Huang, D.-A.; Li, Z.; Li, M.; Liu, G.; Kautz, J.; Alvarez, J.~M.; Zhang, L.; and Yu, Z. 2026{\natexlab{a}}.
\newblock {VideoITG}: Multimodal Video Understanding with Instructed Temporal Grounding.
\newblock In \emph{Proceedings of the IEEE/CVF Conference on Computer Vision and Pattern Recognition}, 24640--24650.

\bibitem[{Wang et~al.(2025)Wang, Li, Yan, He, Yu, Zeng, Wang, Ma, Huang, Gao, Dou, Chen, Wang, Qiao, Wang, and Wang}]{wang2025internvideo25}
Wang, Y.; Li, X.; Yan, Z.; He, Y.; Yu, J.; Zeng, X.; Wang, C.; Ma, C.; Huang, H.; Gao, J.; Dou, M.; Chen, K.; Wang, W.; Qiao, Y.; Wang, Y.; and Wang, L. 2025.
\newblock {InternVideo2.5}: Empowering Video {MLLM}s with Long and Rich Context Modeling.
\newblock arXiv:2501.12386.

\bibitem[{Wang et~al.(2026{\natexlab{b}})Wang, Liu, Gui, Lin, Yang, Liao, Chen, and Zhang}]{wang2025stc}
Wang, Y.; Liu, X.; Gui, X.; Lin, X.; Yang, B.; Liao, C.; Chen, T.; and Zhang, L. 2026{\natexlab{b}}.
\newblock Accelerating Streaming Video Large Language Models via Hierarchical Token Compression.
\newblock In \emph{Proceedings of the IEEE/CVF Conference on Computer Vision and Pattern Recognition (CVPR)}.
\newblock ArXiv:2512.00891.

\bibitem[{Wei et~al.(2025)Wei, Wan, Yu, Wang, Yang, Mao, Zhu, Cai, Wang, Chen et~al.}]{wei2025streamvln}
Wei, M.; Wan, C.; Yu, X.; Wang, T.; Yang, Y.; Mao, X.; Zhu, C.; Cai, W.; Wang, H.; Chen, Y.; et~al. 2025.
\newblock Streamvln: Streaming vision-and-language navigation via slowfast context modeling.
\newblock \emph{arXiv preprint arXiv:2507.05240}.

\bibitem[{Xie et~al.(2026)Xie, He, Wang, Zheng, Ye, and Wu}]{xie2026fluxmem}
Xie, Y.; He, B.; Wang, J.; Zheng, X.; Ye, Z.; and Wu, Z. 2026.
\newblock Fluxmem: Adaptive hierarchical memory for streaming video understanding.
\newblock \emph{arXiv preprint arXiv:2603.02096}.

\bibitem[{Yang et~al.(2025{\natexlab{a}})Yang, Chen, Tian, Wang, Li, Yu, and Jia}]{yang2025visionzip}
Yang, S.; Chen, Y.; Tian, Z.; Wang, C.; Li, J.; Yu, B.; and Jia, J. 2025{\natexlab{a}}.
\newblock VisionZip: Longer is Better but Not Necessary in Vision Language Models.
\newblock In \emph{Proceedings of the IEEE/CVF Conference on Computer Vision and Pattern Recognition (CVPR)}, 19792--19802.

\bibitem[{Yang et~al.(2025{\natexlab{b}})Yang, Zhao, Shukla, Singh, Mishra, Zhang, and Ren}]{yang2025streammem}
Yang, Y.; Zhao, Z.; Shukla, S.~N.; Singh, A.; Mishra, S.~K.; Zhang, L.; and Ren, M. 2025{\natexlab{b}}.
\newblock Streammem: Query-agnostic kv cache memory for streaming video understanding.
\newblock \emph{arXiv preprint arXiv:2508.15717}.

\bibitem[{Yao et~al.(2025)Yao, Li, Wei, Li, Ren, Liu, Ouyang, Wang, Li, Li, Kong, Liu, Zhang, and Sun}]{yao2025timechatonline}
Yao, L.; Li, Y.; Wei, Y.; Li, L.; Ren, S.; Liu, Y.; Ouyang, K.; Wang, L.; Li, S.; Li, S.; Kong, L.; Liu, Q.; Zhang, Y.; and Sun, X. 2025.
\newblock {TimeChat-Online}: 80\% Visual Tokens Are Naturally Redundant in Streaming Videos.
\newblock In \emph{Proceedings of the 33rd ACM International Conference on Multimedia}, 10807--10816.

\bibitem[{Zhang et~al.(2025{\natexlab{a}})Zhang, Li, Cheng, Hu, Yuan, Chen, Leng, Jiang, Zhang, Li, Jin, Zhang, Wang, Bing, and Zhao}]{zhang2025videollama3}
Zhang, B.; Li, K.; Cheng, Z.; Hu, Z.; Yuan, Y.; Chen, G.; Leng, S.; Jiang, Y.; Zhang, H.; Li, X.; Jin, P.; Zhang, W.; Wang, F.; Bing, L.; and Zhao, D. 2025{\natexlab{a}}.
\newblock {VideoLLaMA} 3: Frontier Multimodal Foundation Models for Image and Video Understanding.
\newblock arXiv:2501.13106.

\bibitem[{Zhang et~al.(2025{\natexlab{b}})Zhang, Fan, Ma, Zheng, Huang, Cheng, Gudovskiy, Okuno, Nakata, Keutzer, and Zhang}]{zhang2025sparsevlm}
Zhang, Y.; Fan, C.-K.; Ma, J.; Zheng, W.; Huang, T.; Cheng, K.; Gudovskiy, D.; Okuno, T.; Nakata, Y.; Keutzer, K.; and Zhang, S. 2025{\natexlab{b}}.
\newblock SparseVLM: Visual Token Sparsification for Efficient Vision-Language Model Inference.
\newblock In \emph{International Conference on Machine Learning (ICML)}.

\bibitem[{Zhang et~al.(2024)Zhang, Wu, Li, Li, Ma, Liu, and Li}]{zhang2024videoinstruction}
Zhang, Y.; Wu, J.; Li, W.; Li, B.; Ma, Z.; Liu, Z.; and Li, C. 2024.
\newblock Video Instruction Tuning with Synthetic Data.
\newblock arXiv:2410.02713.

\bibitem[{Zhou et~al.(2025)Zhou, Shu, Zhao, Wu, Liang, Xiao, Qin, Yang, Xiong, Zhang, Huang, and Liu}]{zhou2025mlvu}
Zhou, J.; Shu, Y.; Zhao, B.; Wu, B.; Liang, Z.; Xiao, S.; Qin, M.; Yang, X.; Xiong, Y.; Zhang, B.; Huang, T.; and Liu, Z. 2025.
\newblock {MLVU}: Benchmarking Multi-Task Long Video Understanding.
\newblock In \emph{Proceedings of the IEEE/CVF Conference on Computer Vision and Pattern Recognition}, 13691--13701.

\end{thebibliography}

\end{document}